%% file: main.tex
\documentclass{opt2022} 

\title[A General-Purpose Optimization Solver for Constrained Machine and Deep Learning]{\texttt{NCVX}: A General-Purpose Optimization Solver for \\ Constrained Machine and Deep Learning}

\optauthor{%
\Name{Buyun Liang$^1$} \Email{liang664@umn.edu}\\
\Name{Tim Mitchell$^2$} \Email{tmitchell@qc.cuny.edu}\\
\Name{Ju Sun$^1$} \Email{jusun@umn.edu }\\
\addr $^1$Computer Science \& Engineering, University of Minnesota, Minneapolis, USA\\ 
$^2$Queens College, City University of New York, New York City, USA}

\usepackage{graphicx}
\newcommand{ \Brac }[1]{\ensuremath {\left\lbrace #1 \right \rbrace}}

\newcommand{ \paren }[1]{ \ensuremath {\left(  #1 \right)} }
\newcommand{\norm}[2]{\left\| #1 \right\|_{#2}}

\usepackage{algorithm}
\usepackage{algpseudocode}
\usepackage{xcolor}
\usepackage{amsmath,amsfonts,amssymb, bm}
\usepackage{multicol}
\usepackage{wasysym}

\let\oldhref\href
\renewcommand{\href}[2]{\oldhref{#1}{\bfseries#2}}
\newcommand{\ncvx}{\texttt{NCVX}\ }
\newcommand{\granso}{\texttt{GRANSO}\ }
\newcommand{\pygranso}{\texttt{PyGRANSO}\ }

\newcommand{\T}{\intercal}

\usepackage{listings}
\usepackage{color}

\definecolor{dkgreen}{rgb}{0,0.6,0}
\definecolor{gray}{rgb}{0.5,0.5,0.5}
\definecolor{mauve}{rgb}{0.58,0,0.82}

\input{Utils/math_commands.tex}

\input{Utils/ju_math.tex}
\allowdisplaybreaks

\begin{document}

\maketitle

\begin{abstract}%
Imposing explicit constraints is relatively new but increasingly pressing in deep learning, stimulated by, e.g., trustworthy AI that performs robust optimization over complicated perturbation sets and scientific applications that need to respect physical laws and constraints. However, it can be hard to reliably solve constrained deep learning problems without optimization expertise. The existing deep learning frameworks do not admit constraints. General-purpose optimization packages can handle constraints but do not perform auto-differentiation and have trouble dealing with nonsmoothness. In this paper, we introduce a new software package called \texttt{NCVX}, whose initial release
contains the solver \texttt{PyGRANSO}, a PyTorch-enabled general-purpose optimization package for constrained machine/deep learning problems, \emph{the first of its kind}. \ncvx\,inherits auto-differentiation, GPU acceleration, and tensor variables from PyTorch, and is built on freely available and widely used open-source frameworks. \texttt{NCVX} is available at \url{https://ncvx.org}, with detailed documentation and numerous examples from machine/deep learning and other fields. 

\end{abstract}


\section{Introduction}

Mathematical optimization is an indispensable modeling and computational tool for all science and engineering fields, especially for machine/deep learning. To date, researchers have developed numerous foolproof techniques, user-friendly solvers, and modeling languages for convex (CVX) problems, such as SDPT3~\citep{toh1999sdpt3}, Gurobi~\citep{gurobi}, Cplex~\citep{cplex2009v12}, TFOCS~\citep{becker2011templates}, 
CVX(PY)~\citep{grant2008cvx,diamond2016cvxpy}, AMPL~\citep{gay2015ampl}, YALMIP~\citep{LofbergYALMIP}. These developments have substantially lowered the barrier of CVX optimization for non-experts. However, practical problems, especially from machine/deep learning, are often nonconvex (NCVX), and possibly also constrained (CSTR) and nonsmooth (NSMT).  

There are methods and packages handling NCVX problems in restricted settings: 
PyTorch~\citep{paszke2019pytorch} and TensorFlow~\citep{tensorflow2015-whitepaper} can solve large-scale NCVX, NSMT problems without constraints. CSTR problems can be heuristically turned into penalty forms and solved as unconstrained, 
but this may not produce feasible solutions for the original problems. When the constraints are simple, structured methods such as projected (sub)gradient and Frank-Wolfe~\citep{sra2012optimization} can be used. When the constraints are differentiable manifolds, one can consider manifold optimization methods and packages, e.g., (Py)manopt~\citep{boumal2014manopt,townsend2016pymanopt}, Geomstats~\citep{miolane2020geomstats}, McTorch~\citep{MeghwanshiEtAl2018McTorch}, and Geoopt~\citep{KochurovEtAl2020Geoopt}. For general CSTR problems, KNITRO~\citep{pillo2006large} and IPOPT~\citep{WaechterBiegler2005implementation} implement interior-point methods, while ensmallen~\citep{curtin2021ensmallen} and GENO~\citep{laue2019geno} rely on augmented Lagrangian methods. However, 
moving beyond smooth (SMT) constraints, both of these families of methods, at best, 
handle only special types of NSMT constraints. Finally, packages specialized for machine learning, e.g., scikit-learn~\citep{pedregosa2011scikit}, MLib~\citep{meng2016mllib} and Weka~\citep{witten2005practical}, often use problem-specific solvers that cannot be easily extended to new formulations. 

\section{The \granso and \ncvx packages}

\texttt{GRANSO}\footnote{\url{http://www.timmitchell.com/software/GRANSO/}} is among the first optimization packages that can handle general NCVX, NSMT, CSTR problems~\citep{curtis2017bfgs}:
\begin{align}
\begin{split}
     & \min_{\vx \in \R^n} f(\vx),\;\;\text{s.t. } c_i(\vx) \leq 0,\; \forall\; i \in \mathcal I;\;\; c_i(\vx)=0,\; \forall\; i \in \mathcal E. 
\end{split}
\end{align}
Here, the objective $f$ and constraint functions $c_i$'s are only required to be almost everywhere continuously differentiable. \granso is based on quasi-Newton updating 
with sequential quadratic programming (BFGS-SQP) and has the following advantages: (1) \textbf{unified treatment of NCVX problems}: no need to distinguish CVX vs NCVX and SMT vs NSMT problems, similar to typical nonlinear programming packages;  (2) \textbf{reliable step-size rule}: specialized methods for NSMT problems, such as subgradient and proximal methods, often entail tricky step-size tuning and require the expertise to recognize the structures~\citep{sra2012optimization}, while
\granso chooses step sizes adaptively via a gold-standard line search; (3) \textbf{principled stopping criterion}: \granso stops its iteration by checking a theory-grounded stationarity condition for NMST problems, whereas specialized methods are usually stopped when reaching ad-hoc iteration caps. 

However, \granso users must derive gradients analytically\footnote{\granso is implemented in MATLAB and does not support auto-differentiation, although recent versions of MATLAB have included primitive auto-differentiation functionalities. } and then provide code for these computations,
a process which is often error-prone in machine learning and impractical for deep learning. Furthermore, as part of the MATLAB\ software ecosystem, 
\granso is generally not compatible with popular machine/deep learning frameworks---mostly in Python and R---and users' own existing toolchains.
To overcome these issues and facilitate both high performance and ease of use in machine/deep learning, 
we introduce a new software package called \texttt{NCVX}, whose initial release contains the solver
\texttt{PyGRANSO}, a PyTorch-port of \granso with several new and key features:
(1) auto-differentiation of all gradients, a critical feature to make \pygranso user-friendly; 
(2) support for both CPU and GPU computations for improved hardware acceleration and massive parallelism; 
(3) support for general tensor variables including vectors and matrices, 
   as opposed to the single vector of concatenated optimization variables that \granso uses;  
(4) integrated support for OSQP~\citep{osqp} and other QP solvers 
for respectively computing search directions and the stationarity measure on each iteration. 
OSQP generally outperforms commercial QP solvers in terms of scalability and speed. 
All of these enhancements are crucial for solving large-scale machine/deep learning problems. 
\texttt{NCVX}, licensed under the AGPL V3,  is built entirely on freely available and widely used open-source frameworks; see \url{https://ncvx.org} for documentation and examples.

\section{Usage examples: dictionary learning and neural perceptual attack}
In order to make \ncvx friendly to non-experts, we strive to keep the user input minimal. The user is only required to specify the optimization variables (names and dimensions of variables) and define the objective and constraint functions. Here, we briefly demonstrate the usage of \pygranso solver on a couple of machine/deep learning problems. 


\paragraph{Orthogonal dictionary learning (ODL, \cite{bai2018subgradient})} 

One hopes to find a ``transformation'' $\vq \in \R^n$ to sparsify a data matrix $\mY \in \R^{n \times m}$: 
\begin{align} \label{eq:odl}
        \min_{\vq \in \R^n}\; f(\vq) \doteq 1/m \cdot \norm{\vq^\T \mY}_{1}, \quad \text{s.t.} \; \norm{\vq}_{2} = 1,
\end{align}
where the constraint $\norm{\vq}_{2} = 1$ is to avoid the trivial solution $\vq = \bm 0$. \cref{eq:odl} is NCVX, NSMT, and CSTR: nonsmoothness comes from the objective, and nonconvexity comes from the constraint. Demos 1 \& 2 show the implementations of ODL in \granso and \texttt{PyGRANSO}, respectively. Note that the analytical gradients of the objective and constraint functions are not required in \texttt{PyGRANSO}. 
{
\begin{multicols}{2}
\begin{lstlisting}[language=matlab, caption={\granso for ODL},captionpos=b,label={demo1}]
function[f,fg,ci,cig,ce,ceg]=fn(q)
    f = 1/m*norm(q'*Y, 1);%obj
    fg = 1/m*Y*sign(Y'*q);%obj grad
    ci = [];cig = [];%no ineq constr
    ce = q'*q - 1; % eq constr
    ceg = 2*q; % eq constr grad
end
soln = granso(n,fn);
\end{lstlisting}
\columnbreak

\begin{lstlisting}[caption={\pygranso for ODL},captionpos=b,label={demo2}]
def fn(X_struct):
    q = X_struct.q
    f = 1/m*norm(q.T@Y, p=1) # obj
    ce = pygransoStruct()
    ce.c1 = q.T@q - 1 # eq constr
    return [f,None,ce]
var_in = {"q": [n,1]}# def variable
soln = pygranso(var_in, fn)
\end{lstlisting}
\end{multicols}
}
\paragraph{Neural perceptual attack (NPA, \cite{laidlaw2020perceptual})} The CSTR deep learning problem, NPA, is shown below:
\begin{align}\label{f3}
    \max_{\widetilde{\vx}}\;  \gL \paren{f\paren{\widetilde{\vx}},y},\;\;\text{s.t.}\;\; d\paren{\vx,\widetilde{\vx}} = \norm{\phi\paren{\vx} - \phi \paren{\widetilde{\vx}} }_{2} \leq \epsilon. 
\end{align}
Here, $\vx$ is an input image, and the goal is to find its perturbed version $\widetilde{\vx}$ that is perceptually similar to $\vx$ (encoded by the constraint) but can fool the classifier $f$ (encoded by the objective). The loss $\gL\paren{\cdot,\cdot}$ is the margin loss used in~\citet{laidlaw2020perceptual}. Both $f$ in the objective and $\phi$ in the constraint are deep neural networks with ReLU activations, making both the objective and constraint functions NSMT and NCVX. The $d\paren{\vx,\widetilde{\vx}}$ distance is called the Learned Perceptual Image Patch Similarity (LPIPS)~\citep{laidlaw2020perceptual,zhang2018unreasonable}. Demo 3 is the \pygranso example for solving \cref{f3}. Note that the codes for data loading, model specification, loss function, and LPIPS distance are not included here. It is almost impossible to derive analytical subgradients for \cref{f3}, and thus the auto-differentiation feature in \pygranso is necessary for solving it.

\begin{lstlisting}[caption={\pygranso for NPA},captionpos=b,label={demo3}]
def comb_fn(X_struct): 
    adv_inputs = X_struct.x_tilde 
    f = MarginLoss(model(adv_inputs),labels) # obj
    ci = pygransoStruct()
    ci.c1 = lpips_dists(adv_inputs) - 0.5 # ineq constr. bound eps=0.5
    return [f,ci,None] # No eq constr
var_in = {"x_tilde": list(inputs.shape)} # define variable
soln = pygranso(var_in,comb_fn)
\end{lstlisting}

\input{Sections/Application}

\section{Roadmap}
Although \texttt{NCVX}, with the \pygranso solver, already has many powerful features, we plan to further improve it by adding several major components: (1) \textbf{symmetric rank one (SR1)}: SR1, another major type of quasi-Newton methods, allows less stringent step-size search and tends to help escape from saddle points faster by taking advantage of negative curvature directions~\citep{dauphin2014identifying};  (2) \textbf{stochastic algorithms}: in machine learning, computing with large-scale datasets often involves finite sums with huge number of terms, calling for stochastic algorithms for reduced per-iteration cost and better scalability~\citep{sun2019optimization}; (3) \textbf{conic programming (CP)}:  semidefinite programming and second-order cone programming, special cases of CP, are abundant in machine learning, e.g., kernel machines~\citep{zhang2019conic}; (4) \textbf{minimax optimization (MMO)}:  MMO is an emerging modeling technique in machine learning, e.g., generative adversarial networks (GANs) \citep{goodfellow2020generative} and multi-agent reinforcement learning \citep{jin2020local}.






{\small
\setlength{\bibsep}{0.0pt}
\bibliography{reference}
}


\end{document}

%% file: Utils/math_commands.tex
\usepackage{amsmath,amsfonts,bm}

\def\eqref#1{equation~\ref{#1}}

\def\1{\bm{1}}

\def\eps{{\epsilon}}

\def\vq{{\bm{q}}}

\def\vx{{\bm{x}}}
\def\vy{{\bm{y}}}

\def\mI{{\bm{I}}}

\def\mK{{\bm{K}}}

\def\mV{{\bm{V}}}
\def\mW{{\bm{W}}}

\def\mY{{\bm{Y}}}

\DeclareMathAlphabet{\mathsfit}{\encodingdefault}{\sfdefault}{m}{sl}
\SetMathAlphabet{\mathsfit}{bold}{\encodingdefault}{\sfdefault}{bx}{n}

\def\gB{{\mathcal{B}}}

\def\gL{{\mathcal{L}}}

\newcommand{\R}{\mathbb{R}}



%% file: Utils/ju_math.tex
\usepackage{graphicx,amsfonts,amscd,amssymb,bm,url,color,latexsym,bbm,amsmath}
\usepackage{physics}
\allowdisplaybreaks
\usepackage[capitalize,nameinlink]{cleveref}

\renewcommand{\mathbf}{\boldsymbol}

\newcommand{\mb}{\mathbf}
\newcommand{\mc}{\mathcal}

\DeclareMathOperator{\st}{s.t.}

\usepackage{graphicx}
\let\origtau\tau 
\renewcommand{\tau}{\scalebox{1.44}{$\origtau$}}

%% file: Sections/Application.tex
\section{Constrained deep learning applications}
In this section, we highlight $4$ families of constrained deep learning problems with highly nontrivial, often nonsmooth, constraints. These constraints cannot be easily built into the underlying neural networks. \pygranso\,can directly solve these problems, and detailed codes and tutorials are available on \url{https://ncvx.org/examples}.  

\subsection{Robustness of deep learning models}
In visual recognition, deep neural networks (DNNs) are not robust against perturbations---either adversarially constructed or naturally occurring---that are easily discounted by human perception~\cite{Goodfellow2015ExplainingAH,hendrycks2019benchmarking}. To formalize robustness, one popular way is the \emph{adversarial loss} defined as~\cite{madry2017towards}
\begin{align} \label{eq:robust_loss}
   \begin{split}
        \max_{\mb x'} \ell\paren{\mb y, f_{\mb \theta}(\mb x')} \quad
   \st  \; \mb x' \in \Delta(\mb x)  = \{\mb x' \in [0, 1]^n: d\paren{\mb x, \mb x'} \le \eps\}, 
   \end{split}
\end{align} 
where $f_{\theta}$ is the DNN model, and $\Delta(\mb x)$ 
is the set of allowable perturbations with a radius $\eps$ measured with respect to the metric $d$. Early works assume that $\Delta(\mb x)$ is the $\ell_p$ norm ball intersected with the natural image box, i.e., $\{\mb x' \in [0, 1]^n: \norm{\mb x- \mb x'}_p \le \eps\}$, where $p= 1, 2, \infty$ are popular choices~\cite{madry2017towards,Goodfellow2015ExplainingAH}. To capture visually realistic perturbations, recent work has also modeled nontrivial transformations~\cite{hendrycks2019benchmarking} that use non-$\ell_p$ metrics. For empirical robustness evaluation (RE), solving~\cref{eq:robust_loss} leads to the worst perturbations that could fool $f_{\mb \theta}$.  

An alternative formalism of robustness is the \emph{robustness radius} (or minimum distortion radius), defined as the minimal level of perturbation that can cause $f_{\mb \theta}$ to change its predicted class: 
\begin{align}  
\label{eq:min_distort}
    \min_{\mb x' \in [0, 1]^n}\;  d\paren{\mb x, \mb x'}  \quad \st \; \max_{i \ne y} f_{\mb \theta}^i (\mb x') \ge f_{\mb \theta}^y (\mb x'),  
\end{align} 
where the superscript for $f_{\mb \theta}$ indexes its elements, i.e., the output logits. Solving \cref{eq:min_distort} produces not only a minimally distorted perturbation $\mb x'$ but also a robustness radius, making it another popular choice for RE~\cite{CroceHein2020Minimally, croce2020reliable}. For small and restricted $f_{\mb \theta}$ and selected $d$, \cref{eq:min_distort} can be solved exactly by mixed integer programming~\cite{TjengEtAl2017Evaluating,KatzEtAl2017Reluplex,BunelEtAl2020Branch}. For general $f_{\mb \theta}$ and selected $d$, lower bounds of the robustness radius can be computed~\cite{WengEtAl2018Towards,ZhangEtAl2018Efficient,WengEtAl2018Evaluating,LyuEtAl2020Fastened}. But in general, \cref{eq:min_distort} is heuristically solved via gradient-based methods or iterative linearization~\cite{szegedy2013intriguing,MoosaviDezfooliEtAl2015DeepFool,Hein2017,CarliniWagner2016Towards,rony2019decoupling,CroceHein2020Minimally,PintorEtAl2021Fast}. 

\cref{eq:robust_loss,eq:min_distort} are difficult constrained problems, especially when $d$ is sophisticated---necessary for modeling realistic perturbations~\cite{hendrycks2019benchmarking,EngstromEtAl2019Exploring,WongEtAl2019Wasserstein,LaidlawFeizi2019Functional,HosseiniPoovendran2018Semantic}. Popular exisiting solvers for them rely on explicit projections onto simple sets, and they will not work when $d$ is a non-$\ell_p$ metric. Also, previous work has shown that the quality of the solution using these handcrafted methods is sensitive to key hyperparameters: e.g., step-size schedule and iteration budget~\cite{CarliniEtAl2019Evaluating,croce2020reliable}. With \texttt{PyGRANSO}, we can reliably solve \cref{eq:robust_loss,eq:min_distort} with general $d$'s with minimal hyperparameter tuning~\cite{liang2022optimization}; see \url{https://arxiv.org/pdf/2210.00621} for more details.

\subsection{Neural structural optimization}
Designing physical structures such as bridges, optical devices and airplanes often boils down to structural optimization, a fundamental family of problems in physical science and engineering~\cite{ChristensenKlarbring2008Introduction,hoyer2019neural,chandrasekhar2021auto,chandrasekhar2021tounn}. A primitive version of structural optimization takes the form\footnote{This form is also called \emph{topology optimization}.}:
\begin{align}
\label{eq:str_opt}
    \begin{split}
    \min_{\vx,\bm u}  \bm u^{\T}\mK(\vx)\bm u \quad
    \st  \mK(\vx) \bm u=\bm f,
     \mV(\vx) \le v_0,
     \vx \in \Brac{0,1}^d,
    \end{split}
\end{align}
where $\vx\in \R^d$ is the vectorized binary design variable that indicates where the material should be put to form the structure, $\bm u \in \R^n$ is the displacement vector, $\mb K$ is the global stiffness matrix, and $\bm f$ is the vector of external force. For the constraints, $\mK(\vx) \bm u=\bm f$ encodes the physical laws (e.g., Hooke's law) and is often called the \emph{equilibrium constraint}, and $\mV(\vx) = v_0$ limits the amount of material to be used. 

Recent work~\cite{hoyer2019neural} has introduced the deep image prior (DIP)~\cite{ulyanov2018deep} idea into \cref{eq:str_opt}: $\mb x$ is reparametrized as $\vx = g_{\mb \theta}(\mb \beta)$, where $g_{\mb \theta}$ is a deep neural network parametrized by $\mb \theta$, and $\mb \beta$ is a fixed random input. DIP has shown great promise for solving difficult visual inverse problems (see, e.g., discussions in~\cite{LiEtAl2021Self,WangEtAl2021Early,ZhuangEtAl2022Blind}), and can implicitly promote spatial continuity of the design---which is not explicitly modeled in \cref{eq:str_opt}\footnote{The state-of-the-art methods for structural optimization use smoothing filters during iteration heuristically to encourage spatial continuity; see, e.g., \cite{chandrasekhar2021auto}.}. Also, the overparametrization in DIP tends to ease the global optimization. Relaxing the integer constraint $\vx \in \Brac{0,1}^d$ into a box constraint $\vx \in [0,1]^d$ as is typically done in structural optimization, we arrive at 
\begin{align}
\label{eq:str_opt_dip}
    \begin{split}
    \min_{\mb \theta,\bm u}  \bm u^{\T}\mK(g_{\mb \theta}(\mb \beta))\bm u \quad
    \st  \mK(g_{\mb \theta}(\mb \beta)) \bm u=\bm f,
     \mV(g_{\mb \theta}(\mb \beta)) \le v_0,
     g_{\mb \theta}(\mb \beta) \in [0,1]^d
    \end{split}. 
\end{align}
In order to solve \cref{eq:str_opt_dip}, recent work~\cite{hoyer2019neural} transforms \cref{eq:str_opt_dip} into an unconstrained problem, which includes 1) eliminating $\mb u$ from the physical constraint by solving the linear system $\mK(g_{\mb \theta}(\mb \beta)) \bm u=\bm f$; 2) enforcing the couple of constraints $\mV(g_{\mb \theta}(\mb \beta)) \le v_0$ and $g_{\mb \theta}(\mb \beta) \in [0,1]^d$ via reparametrization, binary search, and implicit differentiation. By contrast, using \pygranso we can directly solve \cref{eq:str_opt_dip} without any of the problem-specific tricks. For realistic design problems, $\mb x$ needs to be discrete-valued (it can have more than $2$ discrete values in multi-material design), and the constraint $\mK(g_{\mb \theta}(\mb \beta)) \bm u=\bm f$ becomes nonlinear---$\mb K$ becomes a nonlinear operator acting on $\mb u$ due to the governing PDEs~\cite{chandrasekhar2021auto,chandrasekhar2021tounn}. \pygranso\,can easily handle these general cases also, whereas the tricks used in~\cite{hoyer2019neural} cannot be generalized. 

\subsection{Orthogonal recurrent neural networks (RNNs)}
The exploding and vanishing gradient issues are common in RNNs, and they could occur whenever the recurrent kernel does not have unit eigenvalues~\cite{lezcano2019cheap,arjovsky2016unitary,harandi2016generalized,bansal2018can,henaff2016recurrent}, i.e., the associated weight matrix is not orthogonal. Recent work proposes imposing orthogonality constraint on the weight matrix directly~\cite{lezcano2019cheap,helfrich2018orthogonal}: 
\begin{align}
\label{eq: orthoRNN}
    \begin{split}
    \min_{\mb \theta} \gL\paren{f_{\mb \theta}(\vx),\vy} \quad
\st \mW_{hh}^{\T}(\mb \theta)\mW_{hh}(\mb \theta) = \mI, \det \mW_{hh}(\mb \theta)=1,
    \end{split}
\end{align} 
where $f_{\mb\theta}$ is the RNN parameterized by $\mb \theta$, and $\mW_{hh}$ is the recurrent kernel of the RNN (i.e., a subvector of $\mb \theta$). Since the constraints define the famous special orthogonal group which is a smooth manifold, manifold optimization methods can be developed to solve \cref{eq: orthoRNN}~\cite{lezcano2019cheap,helfrich2018orthogonal}. But these methods entail heavy mathematics from differential geometry, and hence is impractical for non-experts to implement and build on (there could be additional constraints in real applications). Again, \pygranso\,is able to directly handle the nonlinear constraints, whether the user recognizes or not that the constraint forms the special orthogonal group. 

\subsection{Knowledge-aware machine learning (KAML)}
KAML concerns learning that incorporates prior knowledge, e.g., physical laws as constraints. As an example, it can take the general form of minimizing a loss subject to PDE constraints~\cite{mcclenny2020self,dener2020training}: 
\begin{align}
    \label{eq:pde}
    \begin{split} 
          \min_{u(\mb x)} \; {\mc L(u(\mb x))} 
           \quad \st 
           \begin{cases} 
             f\paren{
            \vx; \frac{\partial u}{\partial x_1},\hdots, \frac{\partial u}{\partial x_d};\frac{\partial^2 u}{\partial x_1 \partial x_1} , \hdots, \frac{\partial^2 u}{\partial x_1 \partial x_d} ;\hdots
            } = 0,\quad \forall\, \vx \in \Omega\\
            \gB\paren{u,\vx}=0,\quad \forall\, \vx \in \partial \Omega
           \end{cases} 
\end{split}. 
\end{align}
Here, $\Omega$ and $\partial \Omega$ denote the domain and its boundary respectively, and the loss $\mc L$ depends on the functional variable $u(\mb x)$ over the domain $\Omega$. The constraints are PDEs, with boundary and/or initial conditions $\gB\paren{\cdot} = 0$. If the loss is a constant, this reduces to solving the PDE problem about $u(\mb x)$. In supervised learning scenarios, the loss could take the form of $1/N \cdot \sum_{i=1}^N \ell(\mb y_i, u(\mb x_i))$, where $\{(\mb x_i, \mb y_i)\}$ is the training set and $u$ is the predictor to be learned from the training set. 

To solve \cref{eq:pde}, one can parametrize $u(\mb x)$ as a neural network, i.e., $u(\mb x; \mb \theta)$. This is natural in modern supervised learning, and is called physics-informed neural networks (PINNs) in the numerical PDE community~\cite{dener2020training,cuomo2022scientific,nandwani2019primal,chen2019deep,huang2021differentiable,lu2021deepxde}. In numerical PDEs, classical methods use finite-difference approximations for all the partial derivatives, whereas the PINN idea directly works with continuous functions that allow partial derivatives to be computed via auto-differentiation. This ``mesh-free" nature of PINNs holds the promise for high-precision solutions even for high-dimensional PDEs. 

The state-of-the-art methods for solving the DNN-parametrized version of \cref{eq:pde} use penalty methods, Lagrangian methods, and augmented Lagrangian methods~\cite{lu2021deepxde,cuomo2022scientific,dener2020training,mcclenny2020self}, which often involve delicate tuning of multiple hyperparameters and could lead to infeasible solutions. By contrast, \texttt{PyGRANSO}\,stops iterations by rigorous check of constraint violation and stationarity.